\documentclass[sigconf,nonacm]{acmart}

\usepackage{amsmath}
\usepackage{array}
\usepackage{booktabs}
\usepackage{graphicx}
\usepackage{placeins}
\usepackage{tikz}
\usetikzlibrary{positioning,arrows.meta,fit,backgrounds}
\usepackage{pgfplots}
\pgfplotsset{compat=1.16}

\setcopyright{none}
\acmConference[Preprint]{Preprint}{}{}
\acmYear{2026}

\newcommand{\gradcorr}{\rho_{\mathrm{grad}}}
\newcommand{\topk}{top-$k$}

\title[Visual Tokens as Side Channels]{The Vision Encoder as a Privacy Boundary:
Visual-Token Side Channels in Encoder-Free Vision-Language Models}

\author{Chenyu Zhou}
\authornote{Corresponding authors.}
\email{zhou.c.76d6@m.isct.ac.jp}
\affiliation{%
  \institution{School of Engineering, Institute of Science Tokyo}
  \city{Tokyo}
  \country{Japan}
}

\author{Qiliang Jiang}
\email{jiangqiliang@zju.edu.cn}
\affiliation{%
  \institution{College of Control Science and Engineering, Zhejiang University}
  \city{Hangzhou}
  \country{China}
}

\author{Shuning Wu}
\email{shuningwu@u.nus.edu}
\affiliation{%
  \institution{Department of Electrical and Computer Engineering, National University of Singapore}
  \city{Singapore}
  \country{Singapore}
}

\author{Xu Zhou}
\authornotemark[1]
\email{zhouxu\_nus@u.nus.edu}
\affiliation{%
  \institution{Department of Electrical and Computer Engineering, National University of Singapore}
  \city{Singapore}
  \country{Singapore}
}


\begin{document}
\flushbottom

\begin{abstract}
A vision encoder compresses image pixels into semantic embeddings, and in doing
so it acts as an implicit privacy boundary between the image and the language
model: the resulting states emphasize semantic content and attenuate the
pixel-local detail needed for exact text recovery. Encoder-free vision-language
models (VLMs) remove this boundary, routing image patches directly into the
language-model token stream. We show that this design choice exposes an
architectural privacy attack surface: the intermediate visual tokens form a
pre-output side channel. Under a token-access adversary, decoders invert the
visual-token streams of two encoder-free VLMs, Gemma4 and Fuyu, into recognizable
image structure and readable held-out access codes (\topk{} exact 21/24 and
22/24, and 42/48 and 46/48 on an independent larger split), while matched
encoder-based controls localize the target region but recover no exact
strings---Qwen3-VL and InternVL on both splits (0/24 and 0/48), and LLaVA-1.5 on
the larger split (0/48). Controlled within-model ablations identify the operative
variable as the spatial sampling fidelity of the visual-token
grid---specifically character-direction sampling density---rather than token or
value count (Fisher $p = 6.52 \times 10^{-7}$, channel projection vs.\ spatial
pooling). The channel is not confined to exported tokens: Gemma4 layer-0
key-value cache tensors are themselves directly invertible ($\gradcorr = 0.4202$
vs.\ $0.0045$ shuffled), placing the side channel on the key-value cache that
production serving stacks persist for decoding efficiency. It survives clutter
and realistic document degradation, transfers zero-shot to public document
images, and resists value-level defenses such as additive noise and
quantization; mitigation must instead reduce the spatial sampling. The vision
encoder thus functions as a privacy boundary whose removal should be treated as a
first-class privacy decision in VLM deployment.
\end{abstract}


\maketitle

\section{Introduction}

Vision-language systems are typically hardened at the output: generated text is
filtered, logged, and access-controlled, while the intermediate visual
representations that produced it are treated as transient implementation
details. This treatment overlooks a more direct privacy target. Intermediate
visual tokens are cached for reuse, logged for debugging and billing, passed to
third-party plugins, and exchanged across split-inference and model-hosting
boundaries---all before any output-side filter runs. If those tokens carry
recoverable image content, then the protection has been bypassed upstream of
the output-side controls operators rely on.

Whether they carry such content is, we argue, a property of the model
\emph{architecture}. Conventional VLMs place a vision encoder between the image
and the language model: the encoder maps pixels to a compact semantic embedding,
discarding low-level appearance in the process. In effect, the encoder functions
as an implicit privacy boundary---the intermediate states it produces carry
semantic content but not pixel-local detail. Encoder-free VLMs remove that
boundary, mapping image patches directly into the language-model token
space~\cite{fuyu2023,diao2024eve,diao2025evev2,li2025breen}. This brings
low-level image evidence much closer to the token stream, and it is an active
architectural direction: Chameleon, EVE/EVEv2, BREEN, and Fuyu pursue token-based
early-fusion and encoder-free multimodal learning, and Tuna-2 reports that pixel
embeddings can match conventional vision encoders for understanding and
generation~\cite{chameleon2024,liu2026tuna2}. As deployments adopt these models,
their visual-token streams become deployment-relevant surfaces.

We organize the paper around a single architectural variable: the
\emph{spatial sampling fidelity} of the visual-token grid. An
encoder-free model produces visual tokens through a near-linear patch-to-token
map, so its token grid is a dense spatial sampling of the image; an
encoder-based model interposes an encoder and pooling that compress that
sampling and replace pixel-local content with semantics. Our central claim is
that this one variable---not token count or raw value budget---governs how much
private content the visual-token side channel exposes.

We instantiate a token-access adversary against four VLMs. Across two
encoder-free models, Gemma4 and Fuyu, the adversary trains decoders that invert
held-out visual-token streams into recognizable image structure far above
shuffled-token controls. Two encoder-based models, Qwen3-VL and InternVL, expose
much weaker structure under the matched attack family; InternVL does so despite a
larger token budget than either encoder-free model. This $2\times2$ separation
localizes the leakage to the encoder-free design rather than to one model.

The side channel goes beyond structure to readable text. Gemma4 recovers
held-out access-code strings at \topk{} exact 21/24, and Fuyu patch-projection
tokens recover 22/24 across independent runs, while matched Qwen3-VL and InternVL
encoder-based controls recover no exact strings under the same
proposal-and-recognition pipeline. We then isolate the mechanism directly.
Spatially pooling Gemma4 tokens eliminates exact code recovery, while projecting
channels but keeping the full spatial grid preserves it; at equal value budgets,
preserving character-direction (horizontal) samples preserves substantially more
readable signal than preserving vertical samples. Raw value count is therefore
not the operative variable---spatial sampling fidelity is. The channel further
persists in early-layer transformer hidden states, survives distractor clutter
and realistic document degradation, transfers zero-shot to public document
images, and resists token-level perturbation defenses because those defenses
leave the spatial sampling untouched.

This paper makes three contributions:
\begin{enumerate}
    \item \textbf{The vision encoder functions as an implicit privacy boundary.}
    Under a token-access threat model, removing it exposes a side channel: two
    structurally distinct encoder-free designs (Gemma4, Fuyu) reconstruct scene
    structure and readable held-out access codes (21/24 and 22/24; 42/48 and
    46/48 on a larger split), while matched encoder-based controls localize the
    code region but recover no exact strings---Qwen3-VL and InternVL (0/24 and
    0/48) and a third control, LLaVA-1.5 (0/48).
    \item \textbf{Spatial sampling fidelity is the causal variable.} Controlled
    within-model ablations---not cross-model comparison---identify the spatial
    sampling fidelity of the visual-token grid, specifically character-direction
    sampling density, as the variable that governs exact recovery, rather than
    token or value count (Fisher $p = 6.52 \times 10^{-7}$).
    \item \textbf{The channel lives on production surfaces and resists
    value-level defenses.} It reaches early-layer hidden states and the Gemma4
    key-value cache that serving stacks persist ($\gradcorr = 0.4202$), survives
    clutter and realistic degradation, transfers zero-shot to public documents,
    and resists value-level noise and quantization---mitigation must instead
    reduce the spatial sampling.
\end{enumerate}

Prior work established that learned features can be inverted; our contribution is
to identify \emph{which} architectural variable governs how much they leak, and
to show that this single variable predicts defense failure, behavior under
realistic degradation, and cross-domain transfer---independent of model identity
or training configuration.

\section{Threat Model and Attack}
\label{sec:threat}

\begin{figure*}[t]
\centering
\includegraphics[width=0.98\textwidth]{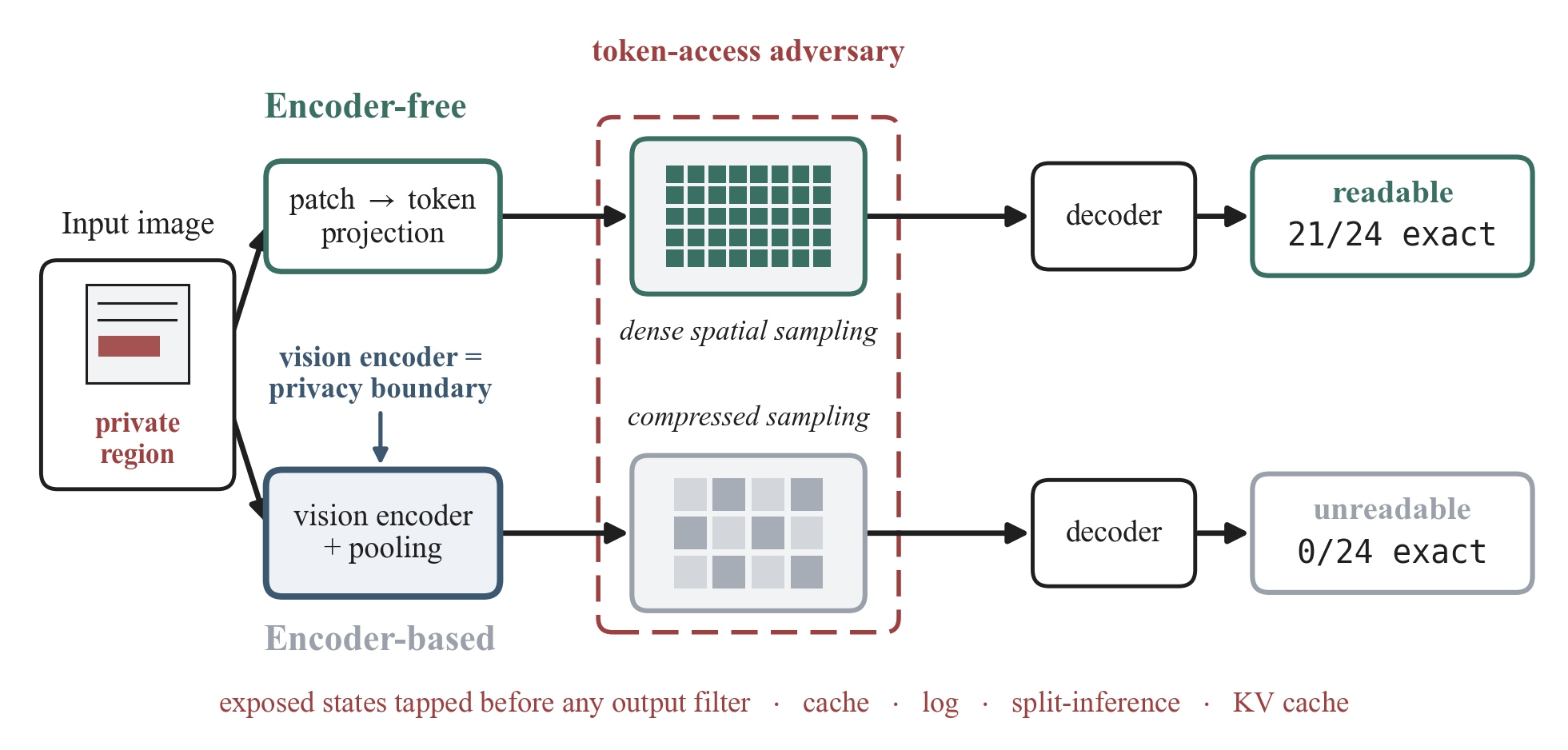}
\caption{Threat model and the architecture split. Both visual paths feed the
same language model, but a token-access adversary observes the visual-token
stream at a cache, log, split-inference, or KV-cache boundary, upstream of any
output filter. Encoder-free tokens are a dense spatial sampling of the image and
invert to readable text; the vision encoder acts as a privacy boundary, so
encoder-based tokens are compressed and invert only to a localized region.}
\Description{A schematic with two lanes from an input image: the encoder-free
lane maps patches through a linear projection to a dense visual-token grid that a
decoder inverts to readable text, while the encoder-based lane passes through a
vision encoder and pooling to compressed semantic tokens that a decoder inverts
only to a localized, unreadable region; a dashed box marks the token states tapped
by a token-access adversary.}
\label{fig:threat}
\end{figure*}

\subsection{Adversary and access points}

The adversary observes intermediate visual tokens or cached visual states rather
than original pixels, and acts upstream of output-side filtering: the visual
state already carries recoverable information independent of what the model
generates (Figure~\ref{fig:threat}). Such access is realistic at several
deployment boundaries that handle visual tokens as ordinary tensors:

\begin{itemize}
    \item \emph{Caching and logging.} Serving stacks cache visual tokens to
    avoid recomputing the vision path across turns, and log intermediate tensors
    for debugging, billing, and quality monitoring. Cached and logged states
    outlive the request and inherit weaker access controls than model outputs.
    \item \emph{Split and remote inference.} Latency- or memory-driven
    deployments split the model across a device and a server, or across
    co-processors, transmitting visual tokens over the boundary in the clear.
    \item \emph{Plugins and multi-tenant hosting.} Tool/plugin interfaces and
    shared hosting expose intermediate representations to components that are
    less trusted than the core model.
\end{itemize}

In each case the adversary's view is the visual-token tensor, not the model's
generated text, so output filters and refusal behavior provide no protection.

These access points share a structural asymmetry that makes the visual-token
stream, not the original image, the natural target. The raw image enters through
a short-lived, access-controlled input-handling path, whereas the visual tokens
and their key-value (KV) cache are persisted and reused as ordinary tensors:
production serving stacks retain the KV cache across turns and inference segments
to avoid recomputation, and do not subject it to the filters applied to outputs. An internal pipeline, plugin, or operations tool with read
access to that cache sees the visual state without ever touching the original
pixels---and, as Section~\ref{sec:breadth} shows, the early KV cache is itself
directly invertible. The exposure is a property of how the architecture stores
and moves visual evidence, not of any single component being compromised.

\subsection{Capabilities and knowledge}

The adversary can collect auxiliary images from a related distribution and knows
which token family is exposed (it attacks each model on that model's native
visual-token grid). It does not access target labels, original pixels, or model
weights at attack time. The adversary uses a \emph{matched attack family} across
all four models: the same decoder architecture, the same
proposal-and-recognition pipeline, and a reconstruction target resolution chosen
to preserve each model's native-grid alignment. The exposed representation is
thus the only variable that changes from model to model.

\subsection{Representation and notation}

We write the exposed visual state of a model as a token tensor
$T \in \mathbb{R}^{H \times W \times C}$, where $(H, W)$ is the spatial token
grid and $C$ is the per-token channel dimension; the \emph{value budget} is the
product $H\,W\,C$. An encoder-free model produces $T$ by a near-linear
patch-to-token map, so the grid $(H, W)$ is a dense spatial sampling of image
patches and each token retains pixel-local appearance. An encoder-based model
interposes a vision encoder and pooling before $T$, which lowers the effective
spatial sampling and replaces pixel-local content with semantic content. The
\emph{spatial sampling fidelity} of $T$ is the combination of its grid density
$(H, W)$ and whether its values still encode pixel-local appearance; this is the
variable the rest of the paper isolates.

The adversary learns a decoder $D$ that maps $T$ to a reconstructed image
$\hat{x} = D(T)$. A shuffled-token control permutes the spatial positions of $T$
before decoding, removing spatial structure while preserving the value
distribution; it is the baseline against which all reconstruction is measured.

\subsection{Models, instantiation, and metrics}

We instantiate the adversary against four VLMs in the main matrix. Two are
encoder-free: Gemma4, whose visual path maps image patches into LLM-space soft
tokens (the visual-token states after the patch-to-LLM projection), and
Fuyu~\cite{fuyu2023}, which linearly projects image patches directly into the
token stream. Two are encoder-based: Qwen3-VL~\cite{qwen3vl}, through its pooler
and pre-pooling vision-encoder states, and InternVL~\cite{internvl2024}, through
its post-projector image tokens. For the text-domain comparison we add a third
encoder-based control, LLaVA-1.5~\cite{llava15}, through its post-projector image
tokens. For each model the adversary attacks the model's native visual-token grid
through the matched proposal-and-recognition pipeline.

We focus on recoverable scene layout and short text strings embedded in images,
with access codes as the labeled text endpoint---a controlled, measurable
representative of short alphanumeric secrets such as PINs and identifiers. Access
codes are a worst-case target---short, alphanumeric, and context-free, with no
language model to fall back on and exact recovery required---so they
conservatively probe the channel's text-carrying capacity. For
text leakage, a proposal selector identifies likely text-bearing regions from
reconstructed-image evidence alone, and a recognizer estimates short-code
readability on labeled access-code pages.

We measure reconstruction structure with gradient correlation ($\gradcorr$),
which compares the edge structure of $\hat{x}$ to the ground-truth image and is
reported against each model's own shuffled-token control. We measure text leakage
as \topk{} exact recovery---the most confident reading over the detector's ranked
proposals, scored against the ground-truth code---with character accuracy (Char)
as the per-character rate and R@1 as proposal recall at rank~1.

\begin{figure*}[t]
    \centering
    \includegraphics[width=0.92\textwidth]{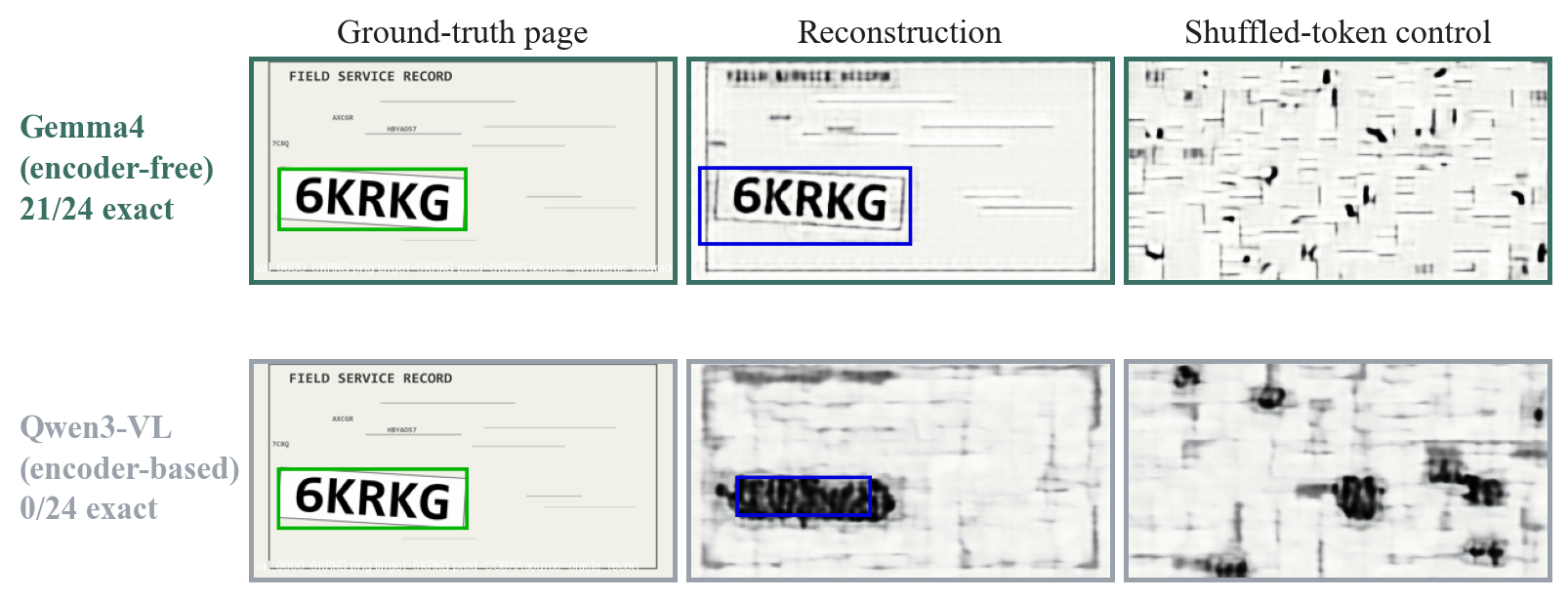}
    \caption{Main text-leakage contrast on the same validation page and attack
    family. Encoder-free Gemma4 visual soft tokens reconstruct a readable
    access-code (\texttt{6KRKG}) and reach \topk{} exact 21/24, while the matched
    encoder-based Qwen3-VL pooler control reconstructs only a dark, unreadable
    region (0/24). Shuffling the token positions destroys the reconstruction for
    both.}
    \Description{A two-row, three-column comparison. Each row shows the
    ground-truth access-code page, the decoder reconstruction, and a
    shuffled-token control. The encoder-free Gemma4 row reconstructs a readable
    access code, while the encoder-based Qwen3-VL row reconstructs only a dark
    unreadable region; the shuffled-token controls are fragmented for both.}
    \label{fig:main-contrast}
\end{figure*}

\section{Experimental Setup}
\label{sec:setup}

To measure the spatial-sampling variable across architectures, we instantiate the
attack with a fixed data, decoder, and evaluation protocol; only the exposed
representation changes between models.

\paragraph{Data.}
Structural inversion uses held-out drone-captured power-line asset-inspection
scenes from InsPLAD~\cite{vieira2023insplad}, split scene-disjointly into
training and validation so that no validation scene appears in training. Text
leakage uses synthetic document pages bearing five-character access codes: codes
are never repeated between training and validation, and pages randomize code
position, scale, slight rotation, and background clutter so that recovery cannot
rely on a fixed template. A realistic-document variant adds multi-font rendering,
forms, tabular noise, code-like distractors, shadows, blur, downsampling, and
JPEG artifacts. Domain transfer uses 22 public receipt, invoice, and bill views.

\paragraph{Exposed states.}
Each model is attacked on its native visual-token state: Gemma4 final visual soft
tokens (the $12\times22\times3840$ states after the patch-to-LLM projection),
Fuyu patch-projection tokens, Qwen3-VL pooler and pre-pooling vision states, and
InternVL post-projector image tokens (Section~\ref{sec:threat}). The
reconstruction target resolution is chosen per model to preserve native-grid
alignment.

\paragraph{Attack pipeline.}
The matched attack family has three stages applied identically across models. A
reconstruction decoder maps the visual-token grid to an image. For text, a
region-proposal stage scores candidate text-bearing windows from the
reconstruction alone, using only code-window size priors estimated from training
pages; validation annotations are used only for evaluation, to score
localization and exact-code recovery on held-out pages. A sequence recognizer,
pretrained on degraded synthetic crops and adapted to reconstructed crops, reads
the proposed region, and the most confident reading over the ranked proposals is
reported as \topk{} exact recovery.

\paragraph{Protocol and controls.}
Structural runs use 128 training and 32 validation images; primary text runs use
96 training and 24 validation pages, with an independent larger $192/48$ split
used to reproduce the text-domain result at greater sample size. Every
reconstruction is paired with a shuffled-token control that permutes token
spatial positions before decoding.
Encoder-based controls receive the same data, decoder family, proposal stage, and
recognizer as the encoder-free models. Defenses are evaluated against an adaptive
attacker that retrains the decoder and recognizer on the defended token stream.

\section{An Encoder-Free Architectural Surface}
\label{sec:architecture}

We first measure the variable at the level of model architecture: do
encoder-free visual tokens retain recoverable image structure where
encoder-based states do not? On held-out drone-captured power-line
asset-inspection scenes from InsPLAD~\cite{vieira2023insplad}, both encoder-free
models reconstruct scene structure far above their shuffled-token controls:
Gemma4 final visual soft tokens reach $\gradcorr = 0.3114$ and Fuyu
patch-projection tokens reach $\gradcorr = 0.6507$, while their shuffled-token
controls stay near zero. The two encoder-based models remain far below this regime under
the matched decoder family: Qwen3-VL pooler states reach $0.0213$, its
finer-resolution pre-pooling vision states ($12 \times 20$ grid) reach $0.0011$,
and InternVL post-projector image tokens reach $0.0959$--$0.1106$ across two
random seeds. Table~\ref{tab:architecture} summarizes the separation.

The separation is not explained by representation size or spatial resolution.
InternVL's token budget ($1{,}048{,}576$ values/image) matches or exceeds both
encoder-free models, yet its $\gradcorr$ stays far below the encoder-free regime.
Qwen3-VL's pre-pooling state exposes a comparable pre-pooling resolution
($12 \times 20$ vs.\ Gemma4's $12 \times 22$), yet is its \emph{least} invertible
state ($0.0011$), so a comparable spatial resolution with encoder compression
yields near-zero structure. This is not an undertrained decoder: the pre-pooling
$\gradcorr$ ($0.0011$) sits at its own shuffled-token control ($0.0010$), so the
state carries no recoverable spatial structure under the matched decoder family.
Within the encoder-based regime the residual signal still tracks grid density---%
InternVL's denser $16 \times 16$ grid gives higher $\gradcorr$ than Qwen3-VL's
$6 \times 10$ pooler ($0.0959$ vs.\ $0.0213$)---but the encoder imposes a ceiling
that no within-class grid density crosses into the encoder-free regime. In the
studied architectures the separation is therefore categorical: the presence of a
vision encoder imposes a readability ceiling that token budget and grid
resolution do not explain. Fuyu replicates the encoder-free result
on an independent architecture, patch grid, and training recipe, ruling out a
Gemma4-specific artifact. Two independent encoder-free positives and three
independent encoder-based negatives (counting the text-domain control of
Section~\ref{sec:text}) mark the leakage as a property of the encoder-free
architectural class.

\begin{table*}[t]
\centering
\caption{Structural visual-token inversion on InsPLAD scenes separates
encoder-free from encoder-based VLMs. Each model is attacked on its native
visual-token grid; $\gradcorr$ is reported against the model's own shuffled-token
control.}
\label{tab:architecture}
\small
\setlength{\tabcolsep}{7pt}
\renewcommand{\arraystretch}{1.15}
\begin{tabular*}{\textwidth}{@{\extracolsep{\fill}}p{0.15\textwidth}p{0.27\textwidth}p{0.21\textwidth}cc@{}}
\toprule
Family & Model (state) & Token grid & $\gradcorr$ & Shuf. \\
\midrule
Encoder-free & Gemma4 soft tokens & $12 \times 22 \times 3840$ & 0.3114 & 0.0004 \\
Encoder-free & Fuyu patch-proj.\ tokens & $9 \times 16 \times 4096$ & 0.6507 & 0.0147 \\
Encoder-based & Qwen3-VL pooler & $6 \times 10 \times 4096$ & 0.0213 & 0.0020 \\
Encoder-based & Qwen3-VL pre-pooling & $12 \times 20 \times 1152$ & 0.0011 & 0.0010 \\
Encoder-based & InternVL projected & $16 \times 16 \times 4096$ & 0.0959--0.1106 & 0.0003--0.0020 \\
\bottomrule
\end{tabular*}
\end{table*}

\section{From Structure to Readable Text}
\label{sec:text}

Structural inversion already crosses a privacy boundary, but the sharpest
demonstration is recovering text that a deployment treats as secret. We move the
same variable to a readable-text endpoint: access-code pages with held-out codes.
Figure~\ref{fig:main-contrast} shows the primary contrast on the same page:
Gemma4 visual soft tokens reconstruct a readable \texttt{6KRKG} crop, while the
matched Qwen3-VL pooler control localizes the target region but leaves it
unreadable. Aggregated over
held-out pages, Gemma4 full-grid text leakage reaches \topk{} exact recovery
21/24 and character accuracy 0.9667. Fuyu independently validates the
encoder-free text result: its patch-projection tokens recover 22/24 exact strings
with character accuracy 0.9833 across independent runs.

The matched encoder-based controls recover no exact strings: Qwen3-VL and
InternVL both reach 0/24 (Table~\ref{tab:text-leakage}). The InternVL row
sharpens the picture into a localization-versus-readability split. InternVL
localizes the target region (proposal R@1 1.0000) at coarse resolution but
recovers no readable string: encoder compression preserves coarse topology while
stripping character-stroke detail (character accuracy 0.1833--0.2083). The
encoder-based states thus carry enough signal to point at where the text is, yet
not enough to read it---exactly the behavior the mechanism in
Section~\ref{sec:mechanism} predicts. Under the identical pipeline, encoder-free
visual-token streams yield readable text while the encoder-based controls do not.

An independent larger split (192/48 held-out codes) reproduces the full
$2\times2$ separation at greater sample size (Table~\ref{tab:text-leakage}, lower
block): encoder-free Gemma4 and Fuyu recover 42/48 and 46/48, while both
encoder-based controls recover 0/48, and all four localize the target at R@1
1.0000---the separation is in readability, not localization. A third
encoder-based control reinforces the boundary: on the identical split,
LLaVA-1.5~\cite{llava15} projected tokens localize the code region (R@1 0.9792)
yet recover 0/48 exact strings. The readable-text split thus holds across three
independent encoder-based controls: the attack localizes the text (R@1 near 1.0
for all three) but recovers no exact string (0/48). Across these models, the
vision encoder is what separates locating text from reading it.

\begin{table*}[t]
\centering
\caption{Access-code text-domain readability under the matched attack family, on
the primary $96/24$ split and an independent larger $192/48$ split. All models
reach proposal R@1 near 1.0 (the decoder localizes the code region for every
model); only the encoder-free streams convert that localization into exact text.
Ranges for Fuyu and InternVL span two independent seeds.}
\label{tab:text-leakage}
\small
\setlength{\tabcolsep}{14pt}
\renewcommand{\arraystretch}{1.15}
\begin{tabular*}{\textwidth}{@{\extracolsep{\fill}}p{0.26\textwidth}ccccc@{}}
\toprule
Condition & n & $\gradcorr$ & R@1 & Exact (\topk{}) & Char \\
\midrule
Gemma4 full-grid text & 24 & 0.7699 & 1.0000 & 21/24 & 0.9667 \\
Fuyu patch-proj.\ tokens & 24 & 0.8377--0.8558 & 1.0000 & 22/24 & 0.9833 \\
Qwen3-VL pooler & 24 & 0.0586 & 1.0000 & 0/24 & 0.0083 \\
InternVL projected & 24 & 0.2612--0.2725 & 1.0000 & 0/24 & 0.1833--0.2083 \\
\midrule
Gemma4 (larger split) & 48 & 0.8509 & 1.0000 & 42/48 & 0.9625 \\
Fuyu (larger split) & 48 & 0.9123 & 1.0000 & 46/48 & 0.9917 \\
Qwen3-VL (larger split) & 48 & 0.0657 & 1.0000 & 0/48 & 0.0542 \\
InternVL (larger split) & 48 & 0.3599 & 1.0000 & 0/48 & 0.2792 \\
LLaVA-1.5 (larger split) & 48 & 0.0362 & 0.9792 & 0/48 & 0.0458 \\
\bottomrule
\end{tabular*}
\end{table*}

\section{Mechanism: Spatial Token Density}
\label{sec:mechanism}

The architecture split of Sections~\ref{sec:architecture}
and~\ref{sec:text} establishes a correlation; the controls in this section
establish causation, by manipulating the variable inside a single model with
everything else---weights, training data, pipeline---held fixed. We interpret the
controls through a spatial-sampling account. A visual
token grid acts as an implicit spatial sampling of the image, so readable text
recovery requires two conditions jointly:
\begin{itemize}
    \item[($C_1$)] the character-direction sampling density resolves individual
    strokes; and
    \item[($C_2$)] the representation still carries pixel-level local
    information rather than encoder-compressed semantics.
\end{itemize}
The two conditions explain the encoder-based controls without invoking model
identity. Encoder-based states fail $C_2$: Qwen3-VL pre-pooling exposes a denser
$12 \times 20$ grid than its pooler yet is the least invertible state in
Table~\ref{tab:architecture}, because its values encode semantics rather than
pixels. Grid density alone is therefore necessary but not sufficient. Within
Gemma4, by contrast, we can break $C_1$ while holding $C_2$ fixed and watch
readability disappear.

The mechanism has a direct geometric interpretation. Each visual token
summarizes one patch of the image, so the token grid
samples the page at a fixed spatial rate. A character is resolved only when enough
samples fall across the direction in which its strokes vary; because text runs
horizontally, that direction is the grid's horizontal axis, and adjacent
characters merge once the horizontal sample spacing approaches the character
width. Pooling along the horizontal axis therefore destroys readable text first,
even when vertical sampling---and hence coarse layout and baseline
structure---remains intact. This is why character accuracy follows horizontal
sampling density while structural $\gradcorr$ does not, and why an encoder that
pools or re-embeds patches into semantic vectors (failing $C_2$) erases
readability regardless of how many values it emits.

\paragraph{Isolating spatial sampling within one model.}
Spatially pooling Gemma4 tokens to a near-Qwen value budget eliminates exact
recovery (0/24) while preserving coarse structure ($\gradcorr = 0.3111$), showing
that spatial grid resolution---not channel count---governs character
readability. Projecting channels while keeping the full $12 \times 22$ spatial
grid, at a comparable value budget, restores readable text: \topk{} exact
recovery 16/24, character accuracy 0.9333 (Table~\ref{tab:budget}). Two
representations with nearly the same value budget thus produce opposite
readability outcomes, separated only by whether the spatial grid is intact.
Two-sided
Fisher's exact tests confirm the channel-projection recovery (16/24) is
significant against both spatial pooling and the Qwen3-VL pooler control (both
0/24; $p = 6.52 \times 10^{-7}$).

\begin{figure}[t]
\centering
\begin{tikzpicture}
\begin{axis}[
  width=\columnwidth, height=5.8cm,
  ybar, bar width=12pt,
  every axis plot/.append style={bar shift=0pt},
  ymin=0, ymax=1.2,
  xmin=0.4, xmax=7.6,
  ylabel={Character accuracy},
  ytick={0,0.25,0.5,0.75,1.0},
  xtick={1,2,3,4,5,6,7},
  xticklabels={{12$\times$22\\1.01M},{12$\times$22c\\304k},{6$\times$22\\507k},{12$\times$11\\507k},{4$\times$22\\338k},{8$\times$11\\338k},{6$\times$11\\253k}},
  x tick label style={font=\scriptsize, align=center},
  yticklabel style={font=\scriptsize},
  ylabel style={font=\footnotesize},
  legend style={font=\scriptsize, at={(0.5,1.24)}, anchor=north, legend columns=1, draw=none},
  nodes near coords, nodes near coords style={font=\tiny, /pgf/number format/fixed, /pgf/number format/precision=2},
  every node near coord/.append style={anchor=west, rotate=90, black},
]
\addplot+[draw=black, fill=black!55] coordinates {(1,0.9667) (2,0.9333) (3,0.30) (5,0.1917)};
\addplot+[draw=black, fill=black!12] coordinates {(4,0.20) (6,0.0167) (7,0.075)};
\legend{full horizontal sampling ($W{=}22$),halved horizontal sampling ($W{=}11$)}
\end{axis}
\end{tikzpicture}
\caption{Character accuracy across the budget controls of
Table~\ref{tab:budget}; each bar is labeled by its grid and value budget
($c$: channel projection). At both equal-budget pairs ($507$k: $6{\times}22$
vs.\ $12{\times}11$; $338$k: $4{\times}22$ vs.\ $8{\times}11$) the configuration
that preserves horizontal character-direction samples ($W{=}22$) recovers more
text, and channel projection at $304$k values exceeds the larger $6{\times}22$
and $4{\times}22$ grids. Value budget does not order the results; preserving the
full spatial grid does, with horizontal samples as the equal-budget
discriminator.}
\Description{A bar chart of character accuracy for seven budget-control grid
configurations. Bars that preserve the full horizontal sampling (width 22) are
tall for the full and channel-projection grids and moderate for the wide grids,
while bars with halved horizontal sampling (width 11) are low; at each equal
value budget the wider grid recovers more text than the taller grid.}
\label{fig:doseresp}
\end{figure}
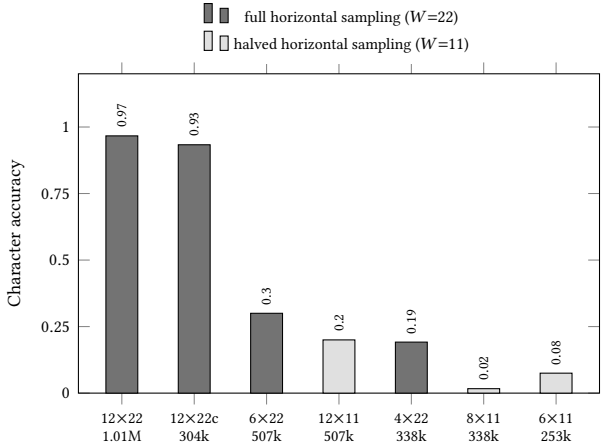

\paragraph{Value count is not the axis.}
Keeping the full grid is what matters even against raw value count: halving the
grid to $12 \times 11$ carries \emph{more} raw values ($506{,}880$) than the
channel-projection control ($304{,}128$) yet drops exact recovery to 0/24. The
$\gradcorr$ column of Table~\ref{tab:budget} makes the dose-response explicit:
structural reconstruction degrades gracefully as the grid coarsens, while exact
recovery has already collapsed---the two metrics dissociate, because exact
recovery depends on resolving strokes and coarse structure does not.
Figure~\ref{fig:doseresp} plots character accuracy across these controls and
shows that the value budget does not order the results.

\begin{figure*}[t]
    \centering
    \includegraphics[width=0.94\textwidth]{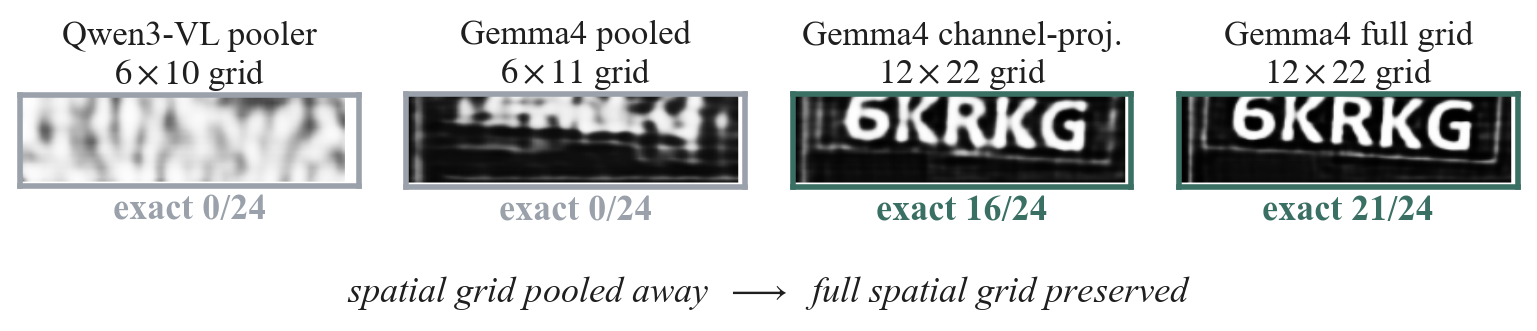}
    \caption{Spatial-density mechanism, reconstruction crops on the same page.
    Qwen3-VL pooling and Gemma4 spatial pooling reduce the spatial grid and leave
    the code unreadable (0/24); Gemma4 channel projection keeps the full
    $12\times22$ spatial grid at a much lower value budget and restores readable
    text (16/24), as does the full grid (21/24). Readable recovery returns when
    the spatial grid is preserved, not when the value budget is.}
    \Description{Four reconstruction crops of the same access code. The Qwen3-VL
    pooler and Gemma4 spatially pooled crops are blurry and unreadable, while the
    Gemma4 channel-projection and full-grid crops show a readable access code.}
    \label{fig:spatial-density}
\end{figure*}

\paragraph{The axis is character-direction sampling.}
The relevant grid direction is the one along which characters extend. Each
equal-budget pair also holds the patch count constant---$6\times22$ and
$12\times11$ both contain 132 patches, $4\times22$ and $8\times11$ both contain
88---so the only free variable within a pair is the horizontal-to-vertical
allocation. At that fixed budget and patch count, preserving horizontal samples
preserves more readable signal than preserving vertical samples: a $6 \times 22$
grid reaches character accuracy 0.3000 versus 0.2000 for a $12 \times 11$ grid
(both $506{,}880$ values), and a $4 \times 22$ grid reaches 0.1917 versus 0.0167
for an $8 \times 11$ grid (both $337{,}920$ values). The taller $12 \times 11$ grid has the
\emph{higher} structural $\gradcorr$ ($0.5127$ vs $0.4678$) yet the
\emph{lower} character accuracy: overall structural fidelity and readable-text
fidelity come apart, and character accuracy tracks character-direction sampling
density specifically (Figure~\ref{fig:spatial-density}). Exact recovery, in turn,
requires the native unpooled spatial grid: channel reduction is tolerated (16/24),
but spatial pooling is not (0/24). Fuyu's independent 22/24
exact recovery on a $9 \times 16$ patch grid is consistent with this account: its
encoder-free patch projection preserves a native spatial grid (satisfying $C_1$)
without encoder compression (satisfying $C_2$).

\begin{table*}[t]
\centering
\caption{Budget controls isolate spatial sampling and character-direction
density. Equal-value pairs share a value budget but differ in grid shape. The
$\gradcorr$ column shows structure degrading gracefully while exact recovery
collapses once the spatial grid is reduced (Fisher's exact test in text).}
\label{tab:budget}
\small
\setlength{\tabcolsep}{14pt}
\renewcommand{\arraystretch}{1.15}
\begin{tabular*}{\textwidth}{@{\extracolsep{\fill}}p{0.24\textwidth}p{0.18\textwidth}cccc@{}}
\toprule
Condition & Grid & Budget & $\gradcorr$ & Exact & Char \\
\midrule
Full grid & $12\times22\times3840$ & 1{,}013{,}760 & 0.7699 & 21/24 & 0.9667 \\
Channel projection & $12\times22\times1152$ & 304{,}128 & 0.5764 & 16/24 & 0.9333 \\
Spatial pooling & $6\times11\times3840$ & 253{,}440 & 0.3111 & 0/24 & 0.0750 \\
Wide grid ($6\times22$) & $6\times22\times3840$ & 506{,}880 & 0.4678 & 0/24 & 0.3000 \\
Tall grid ($12\times11$) & $12\times11\times3840$ & 506{,}880 & 0.5127 & 0/24 & 0.2000 \\
Wide grid ($4\times22$) & $4\times22\times3840$ & 337{,}920 & 0.3314 & 0/24 & 0.1917 \\
Tall grid ($8\times11$) & $8\times11\times3840$ & 337{,}920 & 0.2477 & 0/24 & 0.0167 \\
\bottomrule
\end{tabular*}
\end{table*}

\section{Breadth of the Attack Surface}
\label{sec:breadth}

The same variable governs the channel beyond clean, single-target pages. We track
it along four deployment-relevant axes: representation \emph{depth} inside the
transformer, scene \emph{clutter}, document \emph{degradation}, and \emph{domain
transfer} to real documents. In each case the channel persists exactly as far as
the spatial sampling does. Table~\ref{tab:breadth} collects the results.

\paragraph{Depth: early hidden states and the KV cache.}
The channel persists wherever the early layers still carry the pixel-local
spatial sampling. Gemma4 residual-stream activations at image-token positions
remain invertible in early layers before deeper layers attenuate the pixel-level
signal: layer~0 reaches $\gradcorr = 0.3141$ (shuffle 0.0027), layer~6 reaches
0.2319 (0.0015), and layer~12 reaches 0.0962, with deeper layers continuing to
attenuate. The same structure survives into the model's key-value cache: decoding
Gemma4 layer-0 key/value tensors at image-token positions reaches
$\gradcorr = 0.4202$ (and 0.1656 and 0.1530 at layers~6 and~12). Because the KV
cache is precisely what serving stacks persist and transmit to accelerate
decoding, this places the side channel on a routinely retained surface, not only
on the exported token tensor.

\paragraph{Clutter: target-specific recovery under distractors.}
Clutter changes the page content but not the spatial sampling of the tokens, so
the channel stays target-specific. With the target code embedded among visually
similar distractor codes and labels, Gemma4 reaches \topk{} exact recovery 15/24
and character accuracy 0.9000, the decoder signal remains high
($\gradcorr = 0.7283$ vs.\ shuffle 0.0113), and the automatic proposal selector
never chooses a distractor over the target (0/24). The remaining errors are
character-level recognition failures, not distractor substitutions.

\begin{figure*}[t]
    \centering
    \includegraphics[width=0.9\textwidth]{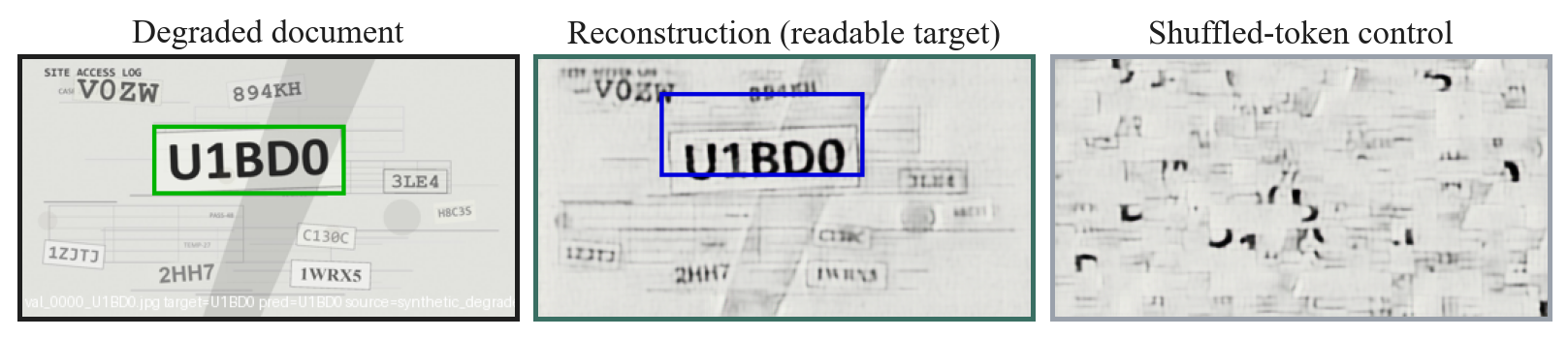}
    \caption{Realistic document leakage. Under multi-font rendering, form clutter,
    shadows, blur, downsampling, and JPEG artifacts, the reconstruction still
    recovers the readable target code (\texttt{U1BD0}), while the shuffled-token
    control is fragmented.}
    \Description{Three panels: a degraded document with the target code, its
    reconstruction with the code still readable, and a fragmented shuffled-token
    control.}
    \label{fig:realism}
\end{figure*}

\paragraph{Degradation: realistic document artifacts.}
Realistic document degradation adds multi-font rendering, forms, tabular noise,
code-like distractors, shadows, blur, downsampling, and JPEG artifacts. These
corrupt the rendered pixels, but the encoder-free path still samples them onto the
same spatial grid, so the decoder signal remains high
($\gradcorr = 0.6895$ vs.\ 0.0044 for shuffled tokens), proposal recall remains
strong (R@1 1.0000), and the target region is reconstructed and localized well
above shuffle (Figure~\ref{fig:realism}). The side channel persists through the
kinds of corruption that real captured documents carry.

\begin{figure*}[t]
    \centering
    \includegraphics[width=0.78\textwidth]{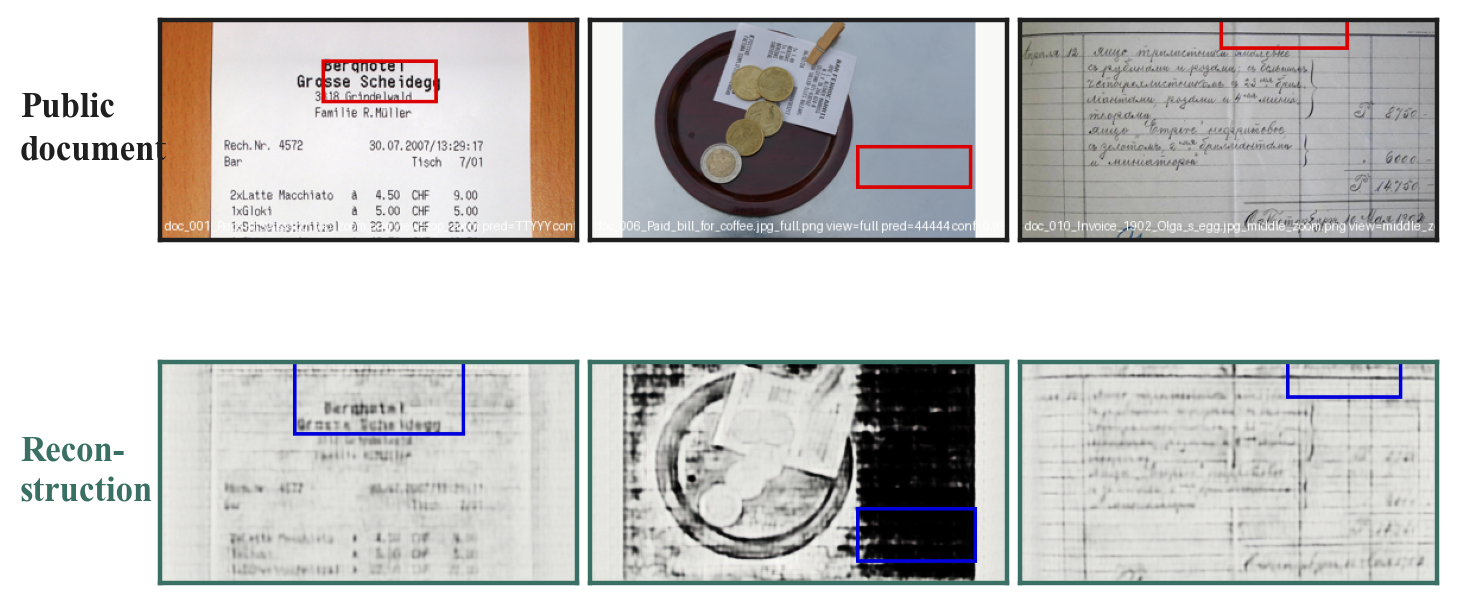}
    \caption{Public real-document transfer. A decoder trained on degraded
    synthetic documents transfers zero-shot to public invoice, receipt, and bill
    images ($\gradcorr = 0.4285$ vs.\ shuffled-token 0.0075); the reconstructions
    preserve document layout and text-region structure with no target-domain
    retraining.}
    \Description{Three public documents (an invoice, a receipt, a handwritten
    bill) each shown next to its reconstruction, which preserves the document
    layout and text-region structure.}
    \label{fig:transfer}
\end{figure*}

\paragraph{Transfer: zero-shot to public documents.}
Real documents share the spatial-sampling structure the decoder learned on
synthetic pages, so the attack transfers without retraining. A decoder trained on
degraded synthetic documents transfers zero-shot to 22 public receipt, invoice,
and bill views, retaining $\gradcorr = 0.4285$ vs.\ $0.0075$ for the
shuffled-token control, with no target-domain retraining. On a public bookkeeping
receipt, the word ``Received'' stays human-readable in the reconstruction, while
the shuffled-token control yields no recognizable word structure
(Figure~\ref{fig:transfer}). The visual-token side channel thus reaches beyond
clean synthetic pages into public document imagery.

\begin{table*}[t]
\centering
\caption{Breadth of the surface: early hidden states, clutter, realistic
degradation, and public-document transfer.}
\label{tab:breadth}
\small
\setlength{\tabcolsep}{12pt}
\renewcommand{\arraystretch}{1.15}
\begin{tabular*}{\textwidth}{@{\extracolsep{\fill}}p{0.26\textwidth}ccp{0.40\textwidth}@{}}
\toprule
Setting & $\gradcorr$ & Shuf. & Evidence \\
\midrule
Hidden state layer 0 & 0.3141 & 0.0027 & early activations invertible \\
Hidden state layer 6 & 0.2319 & 0.0015 & mid-depth still invertible \\
Hidden state layer 12 & 0.0962 & $-$0.0012 & deeper layers attenuate \\
KV cache layer 0 & 0.4202 & 0.0045 & cache tensors invertible \\
Cluttered access codes & 0.7283 & 0.0113 & 15/24 exact; distractor chosen 0/24 \\
Degraded documents & 0.6895 & 0.0044 & readable target region above shuffle \\
Public transfer (n=22) & 0.4285 & 0.0075 & localized text/layout evidence \\
\bottomrule
\end{tabular*}
\end{table*}

\section{Defense Boundary}
\label{sec:defense}

The spatial-sampling account also predicts which defenses can work. It predicts
that a defense closes the channel when it reduces the spatial variable the
channel depends on, whereas a perturbation that leaves the spatial grid and its
pixel-local content intact should be absorbed by an adaptive attacker that
retrains on the perturbed stream.
We test this prediction on Gemma4 full-grid visual soft tokens against an adaptive
attacker that retrains its decoder and recognizer on the defended token stream
(Table~\ref{tab:defense}).

Additive Gaussian noise at $0.10\times$ the token standard deviation leaves the
channel essentially intact ($\gradcorr = 0.7580$, \topk{} exact 18/24, character
accuracy 0.9500), and 3-bit per-feature quantization preserves the reconstruction
signal ($\gradcorr = 0.7823$, \topk{} exact 19/24, character accuracy 0.9583).
Both perturb token \emph{values} while preserving the spatial grid, and both are
absorbed. Reducing the spatial sampling instead---coarsening the grid as in the
budget ablation (Table~\ref{tab:budget})---collapses exact recovery to 0/24. The
defense boundary therefore aligns exactly with the mechanism: effective
mitigation must act on spatial sampling or on the architecture, not on
token-level noise.

\begin{table*}[t]
\centering
\caption{Defenses under an adaptive attacker, on Gemma4 full-grid visual soft
tokens (the Table~\ref{tab:text-leakage} reference row). Token-level noise and
quantization do not close the channel; spatial pooling does (the budget-matched
condition of Table~\ref{tab:budget}).}
\label{tab:defense}
\small
\setlength{\tabcolsep}{16pt}
\renewcommand{\arraystretch}{1.2}
\begin{tabular*}{\textwidth}{@{\extracolsep{\fill}}p{0.34\textwidth}ccc@{}}
\toprule
Defense & $\gradcorr$ & Exact (\topk{}) & Char \\
\midrule
None (reference) & 0.7699 & 21/24 & 0.9667 \\
Gaussian noise $0.10\sigma$ & 0.7580 & 18/24 & 0.9500 \\
3-bit quantization & 0.7823 & 19/24 & 0.9583 \\
Spatial pooling ($6{\times}11$) & 0.3111 & 0/24 & 0.0750 \\
\bottomrule
\end{tabular*}
\end{table*}

\section{Discussion: Deployment Implications}
\label{sec:discussion}

Taken together, the results point to a clear operational conclusion: in
encoder-free VLMs, the visual-token stream is sensitive data, and the effective
protection surface exposed by these experiments is spatial sampling fidelity,
while value-only perturbations leave the channel readable.

\paragraph{Where the boundary sits.}
The exposure is upstream of every output-side control. It appears in exported
visual tokens, in the early-layer hidden states and KV-cache tensors that serving
stacks retain and transmit (Section~\ref{sec:breadth}), and it survives the
value-level perturbations that are most straightforward to deploy
(Section~\ref{sec:defense}). A deployment that filters generated text but caches,
logs, or forwards visual tokens has therefore left its most information-rich
representation exposed. Visual-token streams and their early-layer hidden states should be
subject to the same access-control, retention, and transport protections that
operators already apply to model outputs.

\paragraph{Where mitigation belongs.}
Because the channel is governed by spatial sampling fidelity, mitigation is
effective exactly when it lowers that fidelity. Coarsening the exported spatial
grid removes exact text recovery while value-level noise and quantization do not.
This gives a concrete control surface: operators that must expose intermediate
states can pool or subsample along the spatial grid before the state leaves a
trust boundary, trading reconstruction bandwidth for privacy in a predictable,
dose-responsive way (Table~\ref{tab:budget}). Where utility forbids coarsening,
the protection must be placed structurally---around access to the visual-token
stream itself---rather than in a value-level token perturbation.

\paragraph{A token-access audit.}
The mechanism translates into a concrete audit for an encoder-free deployment.
(i)~Enumerate every component that can observe visual tokens or early
image-token hidden states---caches, logs, split-inference channels, and
plugins---and bring them under the access-control and retention policy used for
outputs. (ii)~Treat the exported spatial grid as the privacy-sensitive quantity:
its density, not its value precision, sets the exposure. (iii)~If intermediate
states must cross a trust boundary, reduce spatial sampling before export and
choose the operating point on the dose-response curve (Table~\ref{tab:budget})
that meets the utility requirement. (iv)~Record the visual-path architecture as a
privacy-relevant attribute, since the encoder-free versus encoder-based choice
determines the exposure regime and whether the boundary carries readable private
content.

\paragraph{Architectural generality.}
The split is not a property of one model. Two encoder-free models leak readable
content and three encoder-based models do not, under one matched attack family,
and the same spatial-sampling variable explains both sides. As encoder-free and
early-fusion designs gain
adoption~\cite{chameleon2024,diao2024eve,diao2025evev2,li2025breen,liu2026tuna2},
the privacy properties of the visual-token boundary should be evaluated as part
of the architectural choice, alongside accuracy and efficiency.

\section{Related Work}
\label{sec:related}

\paragraph{Feature and representation inversion.}
Classic feature inversion established that learned visual representations can
preserve substantial image information: optimization- and network-based
inversions reconstruct recognizable images from deep
features~\cite{mahendran2014inverting,dosovitskiy2015inverting}. We inherit the
inversion methodology but change the object of study from a vision-model feature
map to the visual-token stream of a deployed VLM, and we make the
encoder-free/encoder-based architectural distinction the unit of analysis.

\paragraph{Privacy of embeddings and split inference.}
A second line shows that intermediate states and multimodal embeddings expose
semantic or training-data signal: split-computing activations are vulnerable to
model inversion~\cite{dong2021splitinversion}, and image and multimodal
embeddings leak semantic content or training
data~\cite{chen2026semantic,chen2025leakyclip,jain2025mimic}. Our setting differs
in threat model and endpoint: we attack visual tokens under a token-access
adversary and treat OCR-readable short secrets as a privacy-relevant endpoint,
rather than semantic similarity or membership. Inversion work targets vision
encoders; embedding-privacy work targets semantic embeddings; split-inference
work targets generic activations. None examines the privacy boundary specific to
the encoder-free VLM's pre-output visual-token stream.

\paragraph{Encoder-free and early-fusion architectures.}
These lines converge at the token boundary. Encoder-free, pixel-embedding, and
early-fusion VLM work shows why language-space visual states are becoming
deployment-relevant: Chameleon, EVE/EVEv2, BREEN, and Fuyu pursue token-based
early-fusion and encoder-free multimodal
learning~\cite{chameleon2024,diao2024eve,diao2025evev2,li2025breen,fuyu2023}, and
Tuna-2 shows that pixel embeddings can compete with conventional vision encoders
for multimodal understanding and generation~\cite{liu2026tuna2}. These designs
are evaluated for accuracy and efficiency, and the privacy consequence of routing
pixel-level evidence into the token stream has remained unexamined. Our results
show that this architectural move carries a privacy cost at the visual-token
boundary, and they identify the spatial-sampling property that creates it: two
encoder-free positives, three matched encoder-based negatives, and a single
variable that accounts for both sides.

\section{Conclusion}

The vision encoder functions as an implicit privacy boundary. Removing it, as
encoder-free VLMs do, exposes an architectural visual side channel: Gemma4 and
Fuyu visual tokens reconstruct scene structure and readable short text (21/24 and
22/24; 42/48 and 46/48 on a larger split), while three matched encoder-based
controls---Qwen3-VL, InternVL, and LLaVA-1.5---localize the text but recover
none. Controlled within-model ablations identify the spatial sampling fidelity of
the visual-token grid, specifically character-direction sampling density, as the
variable that governs exact recovery. The surface extends to early-layer hidden
states and the production key-value cache, survives clutter and realistic
document degradation, transfers zero-shot to public documents, and resists
value-level token defenses. The privacy of a VLM's visual-token stream is
therefore set largely by whether pixels pass through a vision encoder and how
densely the visual-token grid samples them, rather than by token-level
parameters---a choice that should be treated as a first-class privacy decision,
with mitigation placed at the spatial-sampling or architectural level rather than
at token-level perturbation.

\bibliographystyle{ACM-Reference-Format}
\bibliography{references}

\section*{Ethical Considerations}

This work studies a pre-output privacy boundary so that VLM deployments can
harden token logging, caching, split inference, and plugin boundaries. The
experiments use controlled synthetic inputs and publicly available document
images. InsPLAD and public-document images measure reconstruction and transfer
behavior. Deployments that expose intermediate visual states to plugins,
split-inference endpoints, or logging pipelines should subject visual-token
streams to the same access-control and retention policies applied to model
outputs.

\section*{Artifact Statement}

The accompanying artifact includes the decoder training scripts, synthetic
document generator, evaluation scripts, figure-generation scripts, and
configuration files needed to reproduce the reported tables. Model weights,
private data, server paths, and non-public images are excluded. Public datasets and model weights are referenced by their official
distribution channels.

\section*{Generative AI Use Statement}

Generative AI tools were used as assistants for code drafting and manuscript
editing. The authors inspected the resulting code, experimental outputs,
citations, and manuscript claims, and remain responsible for the correctness,
originality, and integrity of the manuscript.

\end{document}